\documentclass[letterpaper, 10 pt, conference]{ieeeconf}  

\IEEEoverridecommandlockouts                              

\usepackage{amsmath, amssymb}
\usepackage{graphicx}
\usepackage{subfigure}
\usepackage{xcolor}
\usepackage{algorithm}
\usepackage{algpseudocode}
\usepackage{adjustbox}
\usepackage{subfigure}

\title{\LARGE \bf
Transfer Reinforcement Learning in Heterogeneous Action Spaces using Subgoal Mapping
}

\author{Kavinayan Sivakumar$^{1}$, Yan Zhang$^{2}$, Zachary Bell$^{3}$, Scott Nivison$^{3}$, Michael Zavlanos$^{2}$
\thanks{$^{1}$Kavinayan Sivakumar is with the Electrical and Computer Engineering Department,
        Duke University, NC, USA {\tt\small kps29@duke.edu}
}%
\thanks{$^{2}$Yan Zhang and Michael Zavlanos are with the Mechanical Engineering and Material Sciences Department, Duke University, NC, USA {\tt\small yan.zhang2@duke.edu, mz61@duke.edu}
}%
\thanks{$^{3}$Zachary Bell and Scott Nivison are with the US Air Force, AFRL Division, USA {\tt\small zachary.bell.10@us.af.mil, scott.nivison@us.af.mil}
}%
\thanks{This work is supported in part by AFOSR under award \#FA9550-19-1-0169 and by NSF under award
CNS-1932011.}%
}

\begin{document}

\maketitle
\thispagestyle{empty}
\pagestyle{empty}

\begin{abstract}

In this paper, we consider a transfer reinforcement learning problem involving agents with different action spaces. Specifically, for any new unseen task, the goal is to use a successful demonstration of this task by an expert agent in its action space to enable a learner agent learn an optimal policy in its own different action space with fewer samples than those required if the learner was learning on its own. Existing transfer learning methods across different action spaces either require handcrafted mappings between those action spaces provided by human experts, which can induce bias in the learning procedure, or require the expert agent to share its policy parameters with the learner agent, which does not generalize well to unseen tasks. In this work, we propose a method that learns a subgoal mapping between the expert agent policy and the learner agent policy. Since the expert agent and the learner agent have different action spaces, their optimal policies can have different subgoal trajectories. We learn this subgoal mapping by training a Long Short Term Memory (LSTM) network for a distribution of tasks and then use this mapping to predict the learner subgoal sequence for unseen tasks, thereby improving the speed of learning by biasing the agent's policy towards the predicted learner subgoal sequence. Through numerical experiments, we demonstrate that the proposed learning scheme can effectively find the subgoal mapping underlying the given distribution of tasks. Moreover, letting the learner agent imitate the expert agent's policy with the learnt subgoal mapping can significantly improve the sample efficiency and training time of the learner agent in unseen new tasks.

\end{abstract}

\section{INTRODUCTION}
\label{intro}

Transfer reinforcement learning is an area of active research interest in a variety of applications, spanning speech recognition \cite{generalsequencespeech}, imaging \cite{imagestransfer}, surgery \cite{surgicaltransfericra}, robot manipulation \cite{robotmanipulationtransfericra}, and even gaming \cite{teacherstudentcurriculum, sccurriculumtransfer}. Specifically, as a result of recent advancements in deep learning that have significantly improved computation and performance \cite{deeplearningtransferlearning}, transfer methods are now able to extract important abstract information from the experiences of expert agents performing different tasks and pass them on to other learner agents that have little to no prior knowledge about these tasks, thereby significantly improving the rates at which learner agents learn new tasks.


A common approach to learn an optimal learner policy using the optimal policy of the expert relies on the notion of the Kullback-Leibler (KL) divergence. KL divergence measures the difference between two distributions and, therefore, can be used to capture whether information has been passed on from one agent to another. This approach is popular in teacher-student frameworks, where a teacher expert advises a student target on improving its policy \cite{conditionalteacherstudent, adversarialteacherstudent}. Other popular methods used to transfer knowledge across agents include behavioral cloning \cite{BCcars, BCrobot} and demonstration-based methods \cite{demonstrationfeedbackrobot}, \cite{demonstrationrobottaskspace}. These methods focus on generating the teacher or expert trajectories which are then used to train the source's policy. However, teacher-student and behavioral cloning techniques cannot be directly applied to transfer information across heterogeneous policy functions shaped by different action spaces, as they can lead to unsafe trajectories for the student agent \cite{safetransferprob}. To address this limitation, methods to decompose the policy functions have been proposed so that transfer across different policy structures can be achieved. For example, \cite{modulartransfer} uses a modular approach that decomposes neural network policies into "task-specific" and "robot-specific" modules, the former being shared across robots and the latter shared across all tasks for each agent. Then, the learnt modules are combined to recompose a full policy which can transfer knowledge for a given task. However, meshing of modules together to find an optimal policy can be difficult in complex environments. An alternative transfer learning approach for heterogeneous agents under sparse rewards is proposed in \cite{imithetero} that focuses on the times when the learning agents should imitate the expert versus when they should choose to explore. In practice this method was shown to depend heavily on self-exploration of the learner agent and the benefit of learning from the expert was not clear.

Compared to the transfer learning methods discussed before, mapping approaches are perhaps more intuitive as they capture latent connections between optimal agent trajectories in different action spaces, providing translations between the heterogeneous agents. One such approach is developed in \cite{intertaskmappings} that uses a handcoded mapping between different tasks which have parallels to different action spaces; in both cases having a mapping can help with transferring information. The method in \cite{videogametransferaction} also uses a handcoded mapping, but does not provide competitive experimental results. Although handcoded mappings have been proven to work in some capacity, they are also generally suboptimal and cannot be automatically generated; as a result, additional setup work is required before the transfer learning process can begin \cite{videogametransferaction}. Transfer maps for alignment-based transfer learning have been developed in \cite{alignmenttransferfirst}, \cite{alignmenttransfersafe} to transfer knowledge in robotic manipulation tasks. Specifically, this method uses low dimensional manifolds to represent the source and target datasets of two heterogeneous agents for a particular task and computes a linear mapping between the manifolds. The limitation of a linear mapping is that it may break down in environments with more complex agents. A different approach is proposed in \cite{contexttransfer} that relies on a Markov Decision Process (MDP) homomorphism to transfer knowledge both across different action and state spaces. This approach relies on partial mappings between some features of the source and target tasks and depends on context transferable tasks having the same set of optimal action values across spaces, which is not always guaranteed. Finally, when homogeneous action spaces are considered, mapping methods have been successfully applied for transfer learning, as shown, e.g., in \cite{tensormultitransfer}, that proposes a mapping based on state features to transfer state and action representations between agents in a multi-agent environment, but this approach was not tested for heterogeneous agents.


To address the above limitations of existing solutions, in this paper we learn a mapping between subgoal sequences in the trajectories of an expert and a learner agent. Specifically, we use expert and learner trajectories in a training dataset to learn a Long Short Term Memory (LSTM) network using supervised learning that captures the mapping between expert and learner subgoals, and then use this mapping to predict a learner subgoal sequence for a new task in the testing set. The learner agent's policy is hierarchical, with a high level policy that decides which subgoal state the agent should achieve next and a low level policy that learns the sequence of primitive actions to reach that subgoal. The predicted subgoal sequence for this new task is then used to warm initialize the learner agent's high level policy, after which finding an optimal overall policy is made faster, as the low level policy only needs to optimize agent trajectories in shorter sgments in between subgoals. Our method removes the need for handcoded mappings and does not have any limitations based on linearity on the final mapping.

To our knowledge, most closely related to the method proposed here is the work in \cite{architectureactionspace}, which uses mutual information loss to train embeddings which extract knowledge from the teacher policy to blend with the student policy. However, blending between networks assumes that any representations of knowledge of a state's value from the teacher policy will be useful for the student policy, which is not guaranteed.


The rest of the paper is organized as follows. In Section \ref{formulation}, we formulate the problem of transfer learning across different action spaces and introduce some preliminaries. In Section \ref{method}, we develop our proposed algorithm. Finally, in Section \ref{experiments}, we present experimental results that illustrate the proposed method.

\section{Problem Formulation}
\label{formulation}
Let $s_{expert} \in S_\text{expert}$ and $s_\text{learner} \in S_\text{learner}$ denote the states of an expert and a learner agent, respectively, where $S_{expert}$ is the state space of the expert agent and $S_\text{learner}$ is the state space of the learner agent. Moreover, let $A_\text{expert}$ and $A_\text{learner}$, denote the action spaces of the expert and learner agent that can contain primitive actions, macroactions, or a combination of both. A primitive action is a base action $a \in A$ an agent can perform, while a macroaction is a sequence of primitive actions, $\{a_1, a_2, ..., a_T\}$, where $T$ is the total number of primitive actions in the macroaction. In this paper we assume that the action spaces $A_\text{expert}$ and $A_\text{learner}$ can be different. 

Consider also a distribution of tasks, $d \sim D(p)$, where $d$ is a sampled task parameterized by the task parameter $p$. For example, if $D$ is a distribution of motion planning tasks that require an agent to move to a goal position, $d$ is a task pertaining to a specific end goal position. Here, $p$ may include specifications about the space in which the goal position can be selected, such as length and width of the space. A trajectory associated with task $d$ is defined as a sequence of state action pairs, i.e., $\tau = [(s_0, a_0), (s_1,a_1), ...(s_T, a_T)]$, that accomplishes the desired task, where $T$ is the maximum time horizon. 

Let $r_d(s_t, a_t)$ denote the reward received by the learner agent at state $s_t$ when it takes action $a_t$ in a task $d$. We define a state transition function $s_{t+1} = \rho(s_t, a_t)$. The state value function is defined as $V^\pi(s) = \mathbb{E}_{\psi^\pi}[\sum^\infty_{t=0} \gamma^t r_d(s_t, a_t)|s_0 = s, \pi]$ where $\gamma$ is the discount factor and $\psi^\pi$ is the distribution of states. We parameterize our policy $\pi_\theta$ with policy parameter $\theta$. The objective is to find the optimal policy $\pi^\ast_\theta$ for the learner agent that solves the problem
\begin{equation}
    \underset{\theta}{\max}\hspace{.1cm}J(\theta)
\end{equation}
within a finite time horizon $T$ where $J = \mathbb{E}[V^\pi(s(0)) \vert \psi(s(0)), \pi_\theta]$ is the objective function for the learner agent. In this paper, our goal is to solve the following problem:

\textbf{Problem 1:} (Transfer Reinforcement Learning across heterogeneous action spaces) \textit{Given state spaces $S_{\text{expert}}$ and $S_{\text{learner}}$, action spaces $A_{\text{expert}}$ and $A_{\text{learner}}$ and a data set $W_s \sim D$ containing expert and learner demonstration trajectories $\tau$ for some tasks sampled from the task distribution $D$, use the data in $W_s$ to learn an optimal learner policy $\pi^\ast$ for a new unseen task sampled from $D$, given a new, single expert demonstration of this unseen task, much faster than without transfer of the data in $W_s$.} \\
\begin{figure}
    \centering
    \includegraphics[width=0.2\textwidth]{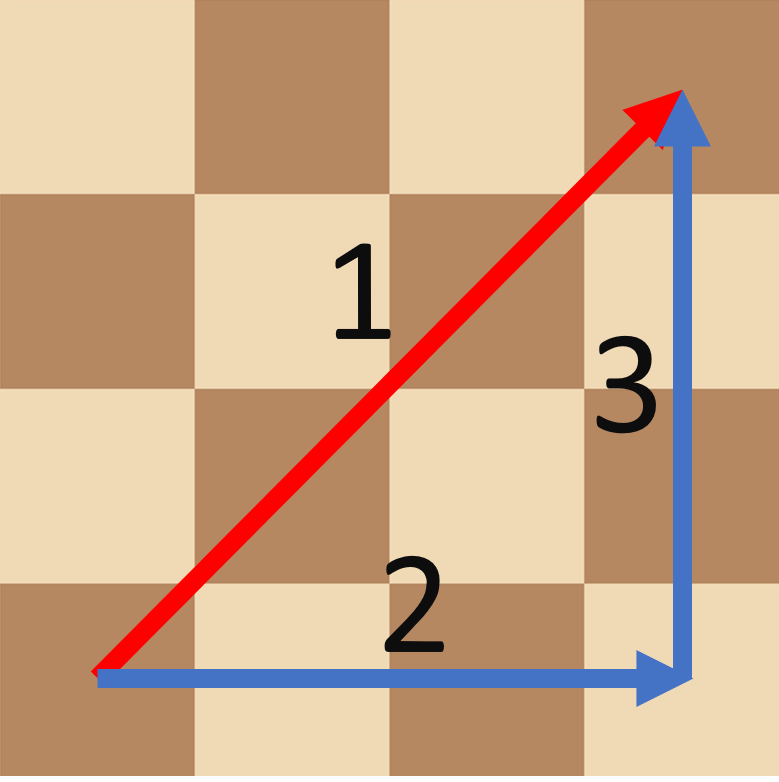}
    \caption{The expert agent needs $2$ actions (blue) to match the displacement made by the learner agent's $1$ action (red). As a result, there is a need to map by sequences, not individual actions.}
    \label{fig:gridexample}
\end{figure}

We note that transferring knowledge directly across action spaces is difficult. To see this, consider two agents $1$ and $2$ that both share state $s$ and whose goal state is $s_f$. Given an action $1_a$ for agent $1$, finding a function $M(s, s_f, 1_a) \rightarrow 2_a$ that outputs the corresponding action for agent $2$ to reach the same goal state $s_f$ from state $s$ is difficult because there is no guarantee for a one-to-one mapping between actions in both action spaces. This is illustrated in Figure~\ref{fig:gridexample}, where two agents, one diagonal and one vertical/horizontal are moving on a grid. This example demonstrates that mapping between different action spaces cannot be done at the primitive action level and it may be required that mappings are designed over sequences of actions. In the example in Figure 1, such a mapping over sequences of actions may take the form $M(s, s_f, \{1_{a, t}\}_{t=0}^T) \rightarrow \{2_{a, t}\}_{t=0}^T$. Still, if this mapping $M$ only maps to the terminal state $s_f$, it may fail to learn latent mappings between agent trajectories on shorter timescales. In this paper, we learn such latent mappings which we define to be mappings between sequences of expert and learner subgoals extracted from their corresponding trajectories. These sequences of subgoals effectively compress sequences of primitive actions and, as a result, avoid the need to map between primitive actions themselves. By extracting sequences of subgoals from agent trajectories, we can learn the high-level behavior of agents whose goal is to achieve a terminal goal state. Moreover, since subsequences of subgoals may appear repeatedly for different terminal goal states, these latent agent behaviors can allow for more robust learning as the mapping is reinforced through behaviors that appear often.
\section{Subgoal Trajectory Mapping Method}
\label{method}
In this section, we first parameterize the mapping function $M$ using a recurrent neural network (RNN) and describe its training procedure. Then, we explain how to use the learned mapping to transfer knowledge from the expert agent to the learner agent so it may learn with fewer samples on unseen tasks within the task distribution.

We assume the expert and learner agents share the same initial state $s_0$ and terminal state $s_f$ and let their corresponding optimal trajectories be given by $\tau_{\text{expert}}$ and $\tau_{\text{learner}}$. Then, we can define the sets of subgoals of the expert and learner agentsf as $G_\text{expert}$ and $G_{\text{learner}}$ respectively, where $g_{\text{expert}} \in G_{\text{expert}}$ and $g_{ \text{learner}} \in G_\text{learner}$ are states in $\tau_{\text{expert}}$ and $\tau_{\text{learner}}$ respectively \cite{Xiao2014AutonomicDO}. Moreover, we define by $g_{t, \text{expert}}$ or $g_{t, \text{learner}}$ a state in $\tau_{\text{expert}}$ or $\tau_{\text{learner}}$ that the expert or learner agent seeks to achieve at time step $t$, respectively. In this paper, we assume that the subgoal sets $G_\text{expert}$ and $G_\text{learner}$ are given. In practice, these subgoal sets can be extracted from the trajectories in the source dataset $W_s$ using unsupervised learning methods, e.g., the algorithm in \cite{shankartemporalskills}. Finally, we also define by $\{g_{t, \text{expert}}\}_{t=0}^T$ and $\{g_{t, \text{learner}}\}_{t=0}^T$ sequences of subgoals of length $T$ drawn from an optimal trajectory $\tau$.

Specifically, in what follows, to solve Problem 1, we first learn a mapping function $M$ that maps the expert agent's subgoal sequences to those of the learner agent through supervised learning; see Section III.A. Using this mapping, we then design a transfer learning algorithm in Section III.B that can reduce the training time for the learner agent in unseen tasks drawn from the distribution $D(p)$, when the learning agent is also provided with a demonstration of this unseen task by the expert agent.
\subsection{Subgoal Sequence Mapping}
We define the subgoal trajectory mapping function as
\begin{equation}
M: (\{g_{t, \text{expert}}\}_{t=0}^T, p) \rightarrow \{g_{t, \text{learner}}\}_{t=0}^T,
\label{eq1}
\end{equation}
where $\{g_{t, \text{expert}}\}_{t=0}^T$ and $\{g_{t, \text{learner}}\}_{t=0}^T$ are defined above and $p$ is the mapping function's context, which can be defined as environment parameters (length and width of a grid environment). Note that the sequence of subgoals returned by the mapping in equation (2) depend on the environmental parameters as the latter dictate the state space and the subgoals predicted by the mapping in (2) are defined on this state space (e.g. space restrictions). Given a set of subgoals used by an expert agent to solve a task with context $p$ in the expert action space, equation (\ref{eq1}) returns another sequence of subgoals that the learner agent can use to solve the same task in its own action space, which is different from the expert's.

To model the mapping function $M$, we adopt a Recurrent Neural Network (RNN), specifically a Long Short-Term Memory (LSTM) network. The network is aware of the total number of subgoals to choose from. In the worst case, these can be all states in the sets $S_\text{expert}$ and $S_\text{learner}$ and we discuss briefly how to define the subgoal set for our specific application in Section IV. Our Keras model \cite{chollet2015keras} encodes the subgoal sequence of the expert and outputs a decoded, predicted subgoal sequence for the learner for a task $d$. For the encoder, we use a bidirectional LSTM network that outputs a fixed-size vector with dimensionality $150$ for the input sequence of subgoals \cite{brownlee2017long}. The decoder is a bidirectional LSTM network of dimensionality $100$ connected to a densely connected softmax layer which outputs a sequence of subgoals for the learner. Bidirectional LSTMs can represent sequences both forward and reverse, allowing for more accurate predictions in sequence to sequence modeling \cite{bidirectionallstm}. Given the softmax probabilities, the model outputs the predicted sequence for the learner. A fully connected layer is used to connect both bidirectional LSTMs together.

Since the sequences of subgoals are temporally related to each other, the positional order of subgoals within the sequence as well as connections need to be captured within the neural network. Then, the mapping function $M$ in equation (2) can be defined by the below general update for LSTMs
\begin{align}
    & h_t = \sigma_h(W_hg_{t, \text{expert}} + U_hh_{t-1} + b_h) \nonumber \\
    & g_{t, \text{learner}} = \sigma_y(W_yh_t + b_y), \nonumber
\end{align}
where $g_{t, \text{expert}}$ and $g_{t, \text{learner}}$ are the $t$-th subgoal of the expert and the learner agent, respectively. The variables $W$, $U$, and $b$ are parameter matrices and vector, $h_t$ is the hidden layer vector, and $\sigma_h$, $\sigma_y$ are activation functions. The parameters of the mapping function $M$ are trained by minimizing the Cross Entropy Loss between the predicted learner agent's subgoal trajectory and the known learner agent's trajectories for the tasks in the training set. This process is further explained in section \ref{experiments}.2.

\begin{algorithm}
\caption{Warm Initializing High Level Learner Policy $\pi_h$ for unseen task $d$ with Supervised Learning}
\textbf{Output:} Warm initialized $\pi_h^{\text{warm}}$
\begin{algorithmic}[1]
\Require Trained mapping $M$, context $p$, expert trajectory $\{g_{t, \text{expert}}\}$ for task $d$, number of iterations E, state $s$, untrained learner high level policy's output weights $w(\pi_h(s))$, noise parameter $v_{\text{noise}}$, subgoal set $G_\text{learner}$
\State $\{g_{t, \text{learner}}\} = M(\{g_{t, \text{expert}}, p\}$
\For {e in E}
\State Initialize empty list $l$ to hold input training pairs
\State $q = \text{random}(0,1)$
\For {g in$\{g_{t, \text{learner}}\}$}
\If {$q \geq v_\text{noise}$}
\State Add ($w$($\pi_h(s)$), g) to $l$
\Else 
\State $g_r = \text{random}$($G_\text{learner}$)
\State Add ($w$($\pi_h(s)$), $g_r$) to $l$
\EndIf
\State $s = g$
\EndFor
\State loss = CrossEntropyLoss($l$)
\State loss.backward()
\EndFor
\end{algorithmic}
\end{algorithm}

\begin{algorithm}
\caption{Learning Optimal Policy $\pi^\ast$ for unseen task $d$}
\textbf{Output:} Optimal policies $\pi_h^{\ast}$ and $\pi_l^\ast$
\begin{algorithmic}[1]
\Require Warm initialized learner high level policy $\pi_h^\text{warm}$, trained learner low level policy $\pi_l$ on seen tasks in distribution $D$, number of training episodes E, number of timesteps T per episode, state $s$, subgoal set $G$, state transition function $\rho$
\For {e in E}
\For {t in T}
\State $g_t = \pi_h^\text{warm}(s_t)$
\If {$g_t \in G$}
\State $a_t = \pi_l(s, g)$
\State $s_{t+1}, r_{t,h}, r_{t,l} = \rho(s_t, a_t)$
\EndIf
\EndFor
\State Update weights for $\pi_h^\text{warm}$ and $\pi_l$ using RL algorithm
\EndFor
\end{algorithmic}
\end{algorithm}

\subsection{Transfer Learning using Subgoal Sequence Mapping}
\label{sec:TF_warm}
Given the predicted sequence of subgoals $\{g_{t, \text{learner}}\}_{t=0}^T$ for the learner, we develop a hierarchical reinforcement learning framework where a high level policy selects the subgoal that the learner agent should pursue next and a low level policy selects primitive actions necessary to meet that selected subgoal. We define the high level policy as
\begin{equation}
\pi_h: s_{t, \text{learner}} \rightarrow g_{t, \text{learner}}
\label{eq2}
\end{equation}
and the low level policy as
\begin{equation}
\pi_l: (s_{t, \text{learner}}, g_{t, \text{learner}}) \rightarrow \{a_{t, \text{learner}}\}.
\label{eq3}
\end{equation}
Specifically, given a new unseen task, the expert agent provides the learner with an expert trajectory of subgoals, which the learner uses to predict its own subgoal trajectory using the trained mapping $M$. Then, using a high level policy \eqref{eq2} that is warm initialized with the predicted subgoal trajectory to output the subgoal sequence needed and applying a low level policy \eqref{eq3} to achieve these subgoals, the learner agent can solve the unseen task using the expert agent's experience, even across different action spaces.

Depending on the training set used to train the mapping $M$, the sequence of predicted learner subgoals may be suboptimal for an unseen task $d$. Therefore, it is desirable that the learner agent can correct for possible errors in the predicted subgoal sequences as it optimizes its policy during learning. Merely sampling from the predicted subgoal trajectory of the mapping $M$ will not necessarily exploit the correctly predicted subgoals more often than the incorrectly predicted subgoals. 

To take advantage of the correctly predicted subgoals in the mapping's predicted subgoal sequence for the learner and help the learner agent's policy converge to the optimal policy faster, we propose to warm initialize the high level policy of the learner with the mapping's predicted subgoal trajectory using supervised learning. This way we can bias the learner's exploration before learning, as seen in Algorithm 1. The input data for supervised learning is the high level policy's output weights associated with the state of the learner agent and the associated label is the predicted subgoal for that state. Along with these training labels, we also incorporate noise into the supervised learning process, which has been shown to improve learning \cite{reed99a}, \cite{trainingwithnoise} and also help the learner's high level policy explore other subgoal options in case the mapping has not predicted the subgoal trajectory for the learner agent accurately. As seen in Algorithm 1, we add noise to the labels in the dataset by setting a probability parameter $v_\text{noise} = [0, 1]$, where for any probability $q \geq v_\text{noise}$, the label is  $g_{t, \text{learner}}$, i.e., the subgoal predicted by the subgoal trajectory returned by the mapping function. For $q < v_\text{noise}$, the label for a state is a random subgoal from the subgoal set $G_{\text{learner}}$. Upon completion of this supervised learning process, the policy weights reflect a bias towards the next subgoal in sequence. We generate data points for the states by incorporating the previously achieved subgoal into the next state. We use cross entropy loss \cite{machinelearningbook} as part of the supervised learning.

After the above warm initialization process for the high-level policy, we also train the low level policy $\pi_l$ in a supervised manner using the trajectories $\tau$ from the learner demonstrations in the seen tasks from the distribution $D$. As opposed to the supervised learning approach to warm initialize the policy $\pi_h$, here we do not incorporate noise as the expert trajectories contain no errors with regards to which action should be taken given a state and desired subgoal. Finally, we train both high-level and low-level policies using any existing RL algorithm as shown in Algorithm 2, (e.g., DPG in \cite{davidsilverdpgpaper} or PPO in \cite{pporlalg}).

\section{Experiments}
\label{experiments}
\subsection{Testing Environment}
In this section, we illustrate the proposed transfer learning method in a chess environment, where the two agents we consider are a dark squared bishop and a knight. Chess has long been a highly regarded game to test reinforcement learning algorithms as it requires calculating strings of moves which can lead to multiple branches \cite{chess}. It is also an ideal experimental environment for reinforcement learning problems with heterogeneous agents, as it has pieces which move and interact very differently from each other. All experiments are implemented using PyTorch \cite{paszke2017automatic} on a Ubuntu system with Nvidia RTX 2080 Ti, while the mapping model $M$ is implemented using Keras \cite{chollet2015keras}. 

\begin{figure}
\centering 
\begin{subfigure}
  \centering
  \includegraphics[width=.45\linewidth]{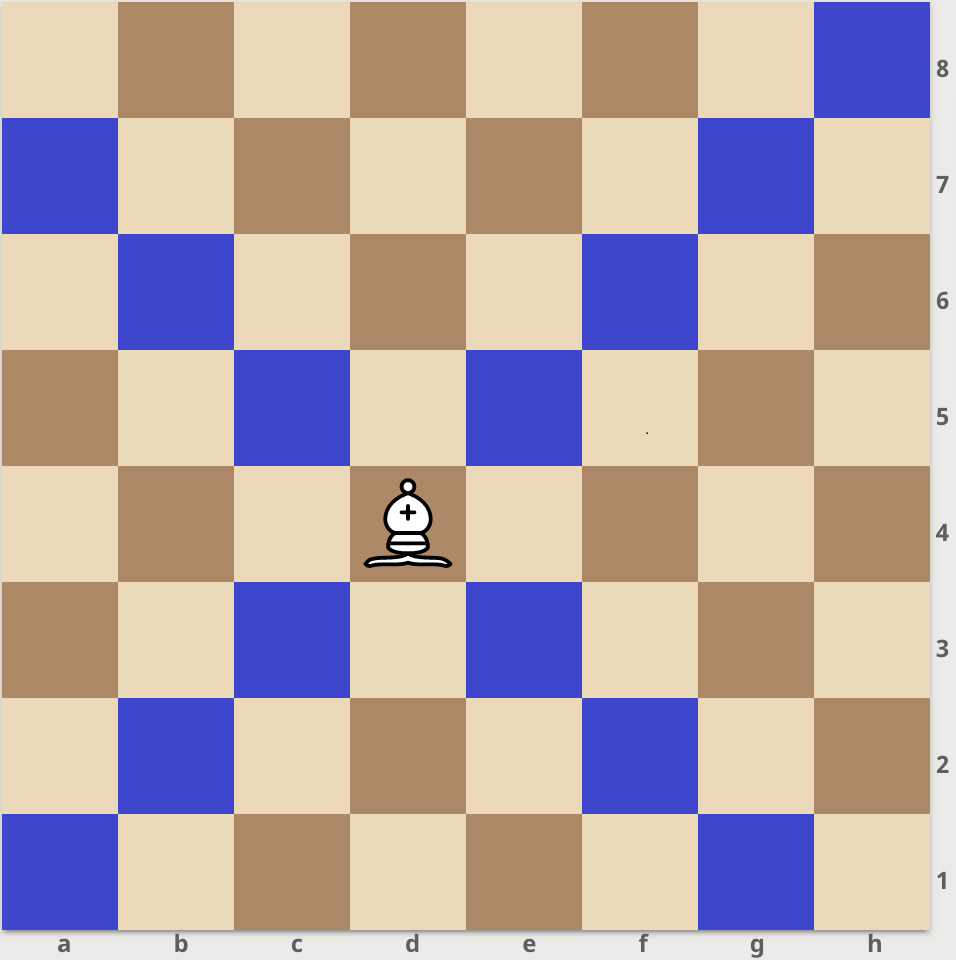}
  \label{fig:sub1}
\end{subfigure}%
\begin{subfigure}
  \centering
  \includegraphics[width=.45\linewidth]{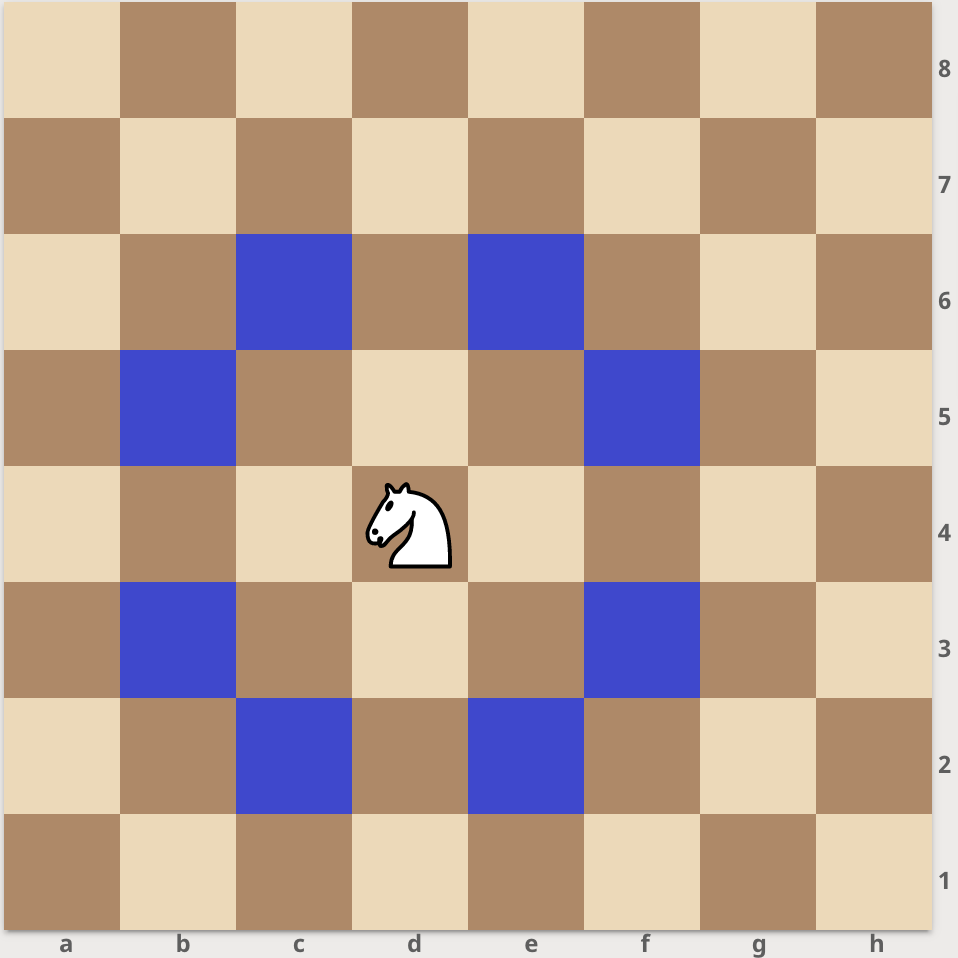}
  \label{fig:sub2}
\end{subfigure}
\caption{The actions a bishop and a knight can take from the center of the board in one step.}
\label{fig:chesspieces}
\end{figure}

Figure~\ref{fig:chesspieces} describes the different action spaces and transitions of the bishop and knight agents respectively. Specifically, each blue square represents the next square these agents can move to. Note that the bishop cannot surpass other pieces if they are on its way. According to Figure~\ref{fig:chesspieces}, these two agents have heterogeneous action spaces and transition functions. 

In the experiments, the bishop or knight agent is located on an $8 \times 8$ board with a number of opposing pawns put in legal positions. Here, we define legal positions as any dark squares not on the first or last row of the chess board, so as to accommodate the dark squared bishop's limitations. 
The goal of the agents is to take as many pawns as possible with the least number of moves. The reward functions for the high level policy $\pi_h$ and the low level policy $\pi_l$ are defined as follows
\[
  r_{t,h}=\begin{cases}
               -1:\text{agent lands on a square that has no pawn,}\\
               10:\text{agent captures a pawn,}
            \end{cases}
\]
\noindent \[
  r_{t, l}=\begin{cases}
               -1:\text{agent does not arrive at desired subgoal,}\\
               10:\text{agent arrives at desired subgoal.}
            \end{cases}
\]

When all pawns are captured or a finite number of timesteps has passed, the episode ends. Specifically, in our experiments, we have $2$ pawns to capture. The set of tasks contains all possible positions of the pairs of pawns in legal positions. The overall task distribution contains $253$ tasks, with the training set of size $228$ and the testing set of size $25$. The tasks in the training and testing sets are randomly allocated. In addition, we assume that we have the optimal sequences of moves of the bishop and the knight agents for the tasks in the training dataset, which are learnt through Djikstra's algorithm. The goal is to transfer the optimal moves of the bishop for the tasks in the testing dataset to help the knight agent learn its optimal moves for the unseen tasks in the testing dataset.


\subsection{LSTM Network Prediction Accuracy}
We train a LSTM network to represent the subgoal sequence mapping function $M$ in \eqref{eq1}. Specifically, the LSTM network receives the embedding of the bishop agent's subgoal trajectory as the input, 
and outputs the probability of selecting each square on the board as a subgoal for the knight agent. The initial hidden layer of the LSTM network, $h_0$, is defined as the starting position of the agent and the positions of the pawns. We use categorical cross entropy as the loss function \cite{machinelearningbook}, which is defined as
\begin{equation}
    L = -\sum_{i=1}^N \mathbb{P}(g_{t, \text{learner}})\text{ log} (\mathbb{P}(\hat{g}_{t, \text{learner}})),
\end{equation}where $N$ is the total number of subgoals in the set $G_\text{learner}$, $\mathbb{P}(g_{t, \text{learner}})$ is the probability of the target subgoal from a trained task at time step $t$, and $\mathbb{P}(\hat{g}_{t, \text{learner}})$ is the probability of the model's output subgoal at time step $t$. The neural network is trained so that the categorical cross entropy loss function is minimized given the bishop and knight subgoal trajectories in the training dataset. 

In order to assess how accurate the mapping $M$ is at predicting subgoal trajectories, we use a metric for machine translation output used in natural language processing (NLP), METEOR \cite{banerjee-lavie-2005-meteor}. Table~\ref{sample-table} shows the METEOR scores across a K-fold cross-validation on the overall task set with $K=10$. Note that the function $M$ is trained with both the bishop and knight trajectories in the training set. Then, the scores in Table~\ref{sample-table} reflect the quality of predicting the subgoal trajectories for the knight agent on tasks in the testing set.
\begin{table}
  \caption{LSTM Prediction Accuracy (Average = 0.5090)}
  \label{sample-table}
  \centering
  \begin{tabular}{ll}
    K-Fold     & METEOR Score    \\
    1 & 0.5089       \\
    2     & 0.4623     \\
    3     & 0.5654        \\
    4 & 0.4357      \\
    5     & 0.4598     \\
  \end{tabular}
  \begin{tabular}{ll}
    K-Fold     & METEOR Score    \\
   6     & 0.6573        \\
    7 & 0.5100       \\
    8     & 0.5271     \\
    9     & 0.4515        \\
    10 & 0.5123 \\
  \end{tabular}
\end{table}
Specifically, if there is one error in the mapping's prediction, the METEOR score is approximately 0.6389, while if there are two errors, the METEOR score is approximately 0.4688. As a result, the average accuracy across the K-folds is between $1$ and $2$ errors in the mapping's prediction. As each subgoal trajectory in the dataset consists of $4$ to $5$ subgoals on average, we can see that the mapping predicts approximately half of the subgoals correctly.

\subsection{Transfer Learning with Warm Initialization}

\begin{figure}
    \centering
    \includegraphics[width=0.25\textwidth]{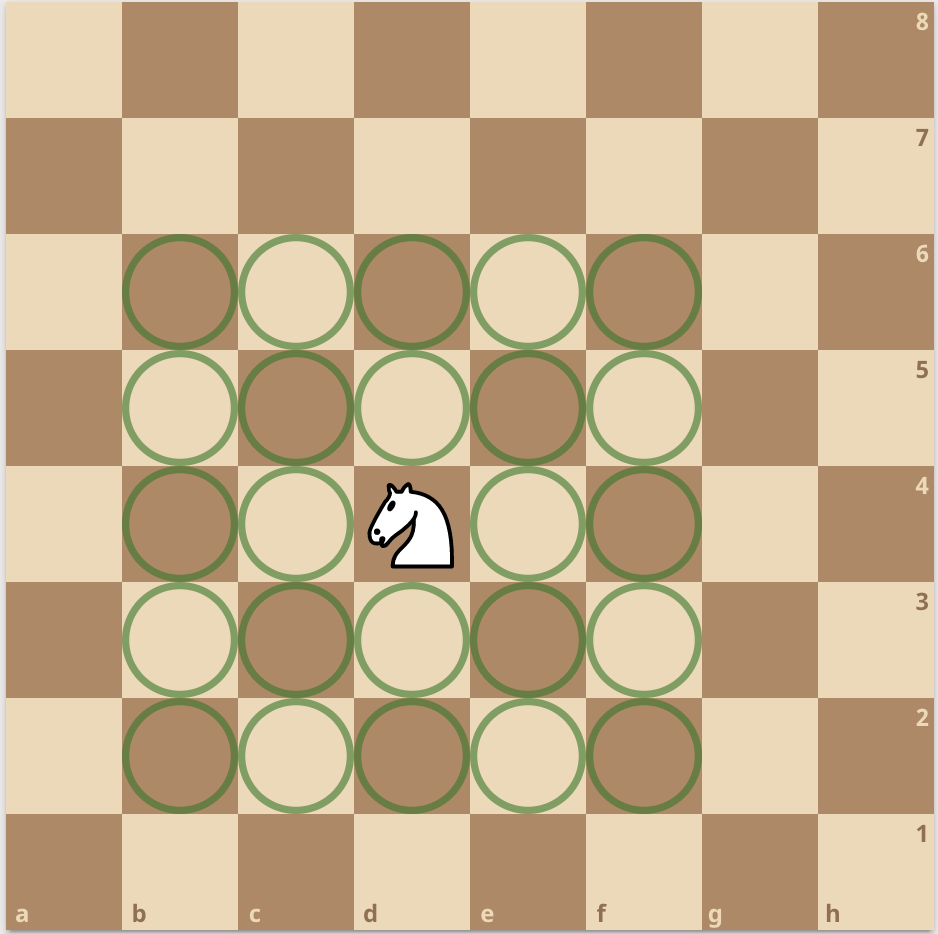}
    \caption{The high level policy subgoal space for a knight agent in the center of the board. The agent can choose any of these subgoals for its next target during the trajectory.}
    \label{fig:chesspieces3}
\end{figure}
In this section, we warm initialize the learner high level policy with the mapping's predicted subgoals as discussed in Section~\ref{sec:TF_warm}. Then, we use deterministic policy gradient (DPG) \cite{davidsilverdpgpaper} to train high level and low level policy functions $\pi_h$ and $\pi_l$. 
For the knight agent, we define its subgoal set at a specific position in Figure~\ref{fig:chesspieces3}. Specifically, the knight can select any square at most $2$ squares away from its current position. This is to ensure that successive subgoals are not too far away from the current position of the knight agent.

\begin{figure}
    \includegraphics[width=0.5\textwidth]{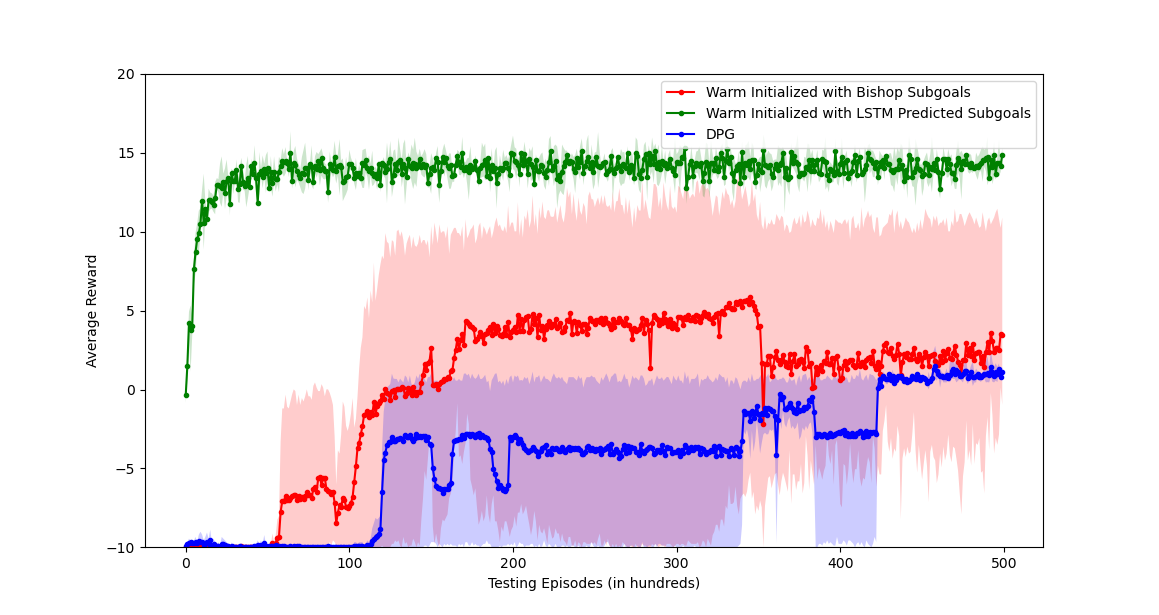}
    \caption{The learning curves for an unseen task where the mapping has correctly predicted the subgoal trajectory for the knight. The policy warm initialized with the mapping's predictions (green) quickly converges to the optimal through DPG as opposed to the policy warm initialized with the bishop's subgoal trajectory (red) and the policy learnt without any warm initialization (blue).}
    \label{fig:results1}
\end{figure}

\begin{figure}
    \includegraphics[width=0.5\textwidth]{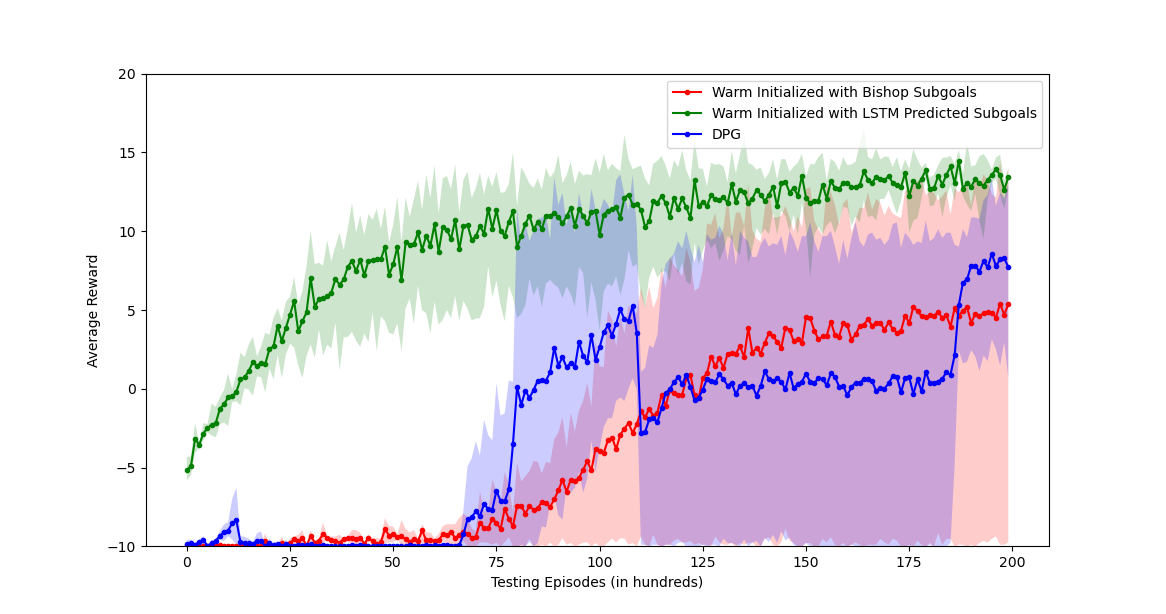}
    \caption{In this unseen task, the mapping predicts one of the subgoals incorrectly. The mapping guided policy (green) still beats both other methods and is able to find an optimal policy, as opposed to policy just using DPG without any warm initialization (blue) and bishop guided policy (red).}
    \label{fig:results3}
\end{figure}

\begin{figure}
     \includegraphics[width=0.5\textwidth]{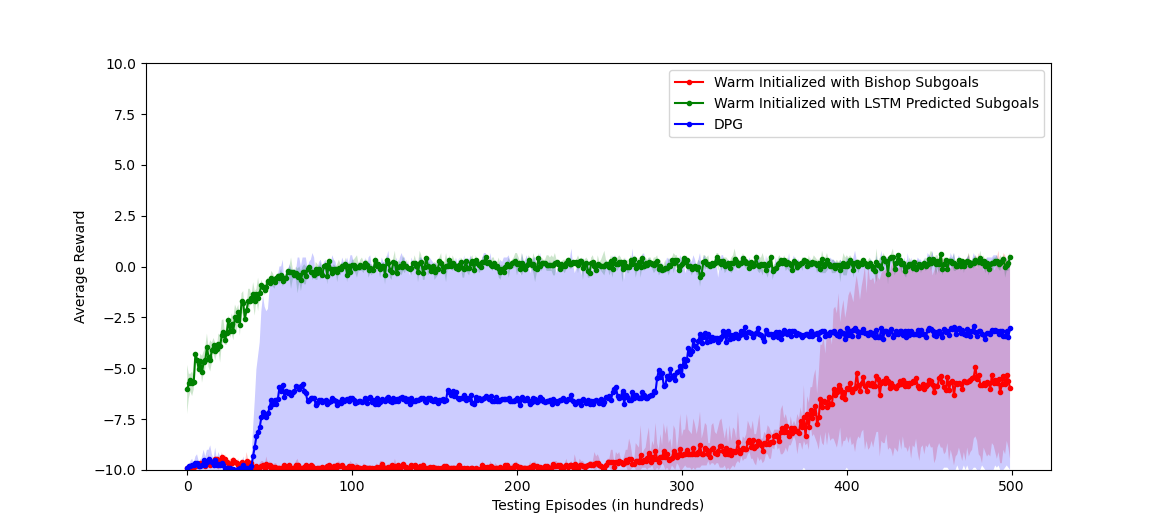}
    \caption{In this unseen task, the mapping fails to predict any of the subgoals correctly, but the policy guided by the mapping (green) still reaches a higher local maximum, which the bishop subgoal guided policy (red) and the policy using just DPG without warm initialization (blue) do not.}
    \label{fig:results4}
\end{figure}

We can split the mapping's performance on a test set into three distinct cases: 1) the mapping predicts the subgoal trajectory without any errors; 2) the mapping predicts the subgoal trajectory with $1$ or $2$ errors; and 3) the mapping predicts the subgoal trajectory with $3$ or more errors. Since most subgoal trajectories of the knight within this task set of $253$ tasks have a length of $4$, having $1$ or $2$ errors may still have significant impact. For each one of the three cases above, we compare the performance of three different approaches: transfer learning using the mapping's suggested subgoals as discussed in Section~\ref{method}; learning the optimal policy using DPG without any transfer; and warm initializing the high level policy $\pi_h$ with the bishop's subgoal trajectory directly and then polishing the policy functions using DPG. 
We train each method for $20$k episodes. Figures~\ref{fig:results1}-\ref{fig:results4} show episode rewards averaged every $100$ episodes. Each curve is averaged over $3$ trials.


In Figure \ref{fig:results1} we present learning curves for a task in the testing set for which the mapping has correctly predicted the sequence of subgoals. The policy that is warm initialized with the subgoal predictions given by the mapping function $M$ reaches its optimal value while the other two methods fail to do so. In Figure \ref{fig:results3}, the mapping only partially predicts the correct subgoal trajectory for the knight in an unseen task. Even so, the policy warm initialized by the partially correct subgoal trajectory is able to achieve its optimal value within $20$k episodes while the other two methods do not. This suggests that the policy function that is warm initialized with partially correct subgoal trajectories can still help high-level exploration and use fewer samples to find correct subgoals to overwrite the incorrect subgoals predicted by the mapping function $M$. Finally, in Figure \ref{fig:results4}, we present the learning curves when the mapping does not predict the subgoals correctly. The policy warm initialized by the mapping is still able to reach a local optimal value much faster than the policy that is trained with DPG without transfer or the policy warm initialized by the bishop subgoal trajectory directly.

We conclude that transferring the expert agent's subgoal trajectory to the learner agent using the method discussed in Section~\ref{method} can help exploration and accelerate the learning process for the learner agent even when the trained mapping function $M$ makes mistakes in predicting subgoals for unseen tasks. Moreover, the proposed method demonstrates significant improvements in learning performance compared to learning from scratch or transferring the subgoal trajectory directly from the other heterogeneous agent.

\section{Conclusion}
In this paper, we studied a transfer reinforcement learning problem across heterogeneous agents, where a recurrent neural network is used to map between subgoal sequences of both agents. The goal of the mapping is to predict the optimal subgoal sequence of the learner agent for an unknown task given the subgoal sequence of the expert. Given this prediction, we warm initialize the high level policy to bias it to pursue the predicted subgoal sequence from the mapping initially as it learns an optimal policy using a standard reinforcement learning algorithm. We demonstrated how to design this mapping using a LSTM recurrent neural network and provided numerical examples showing the effectiveness of our proposed method.

\addtolength{\textheight}{-8.75cm}   



\newpage
\bibliographystyle{IEEEtran}
\bibliography{IEEEabrv, biblio}
\end{document}